\documentclass[letterpaper]{ieeeconf}
\IEEEoverridecommandlockouts
\usepackage{cite}
\usepackage{amsmath,amssymb,amsfonts}
\usepackage{algorithmic}
\usepackage{graphicx}
\usepackage{textcomp}
\usepackage{xcolor}
\usepackage{multicol}
\usepackage[ruled,vlined ]{algorithm2e}
\usepackage{booktabs}
\usepackage{subcaption}
\usepackage[%
hyperfigures=true,%
backref=page,%
pagebackref=true,%
breaklinks=true,%
colorlinks=true,%
citecolor=green,%
linkcolor=blue
]{hyperref}
\usepackage{pbox}
\let\labelindent\relax
\usepackage{enumitem}
\usepackage{verbatim}
\usepackage[bottom=0.75in, top = 0.75in, left =0.75in, right =0.75in]{geometry}
\newtheorem{assumption}{Assumption}

\newcommand{\vthre}[3]{\begin{bmatrix}#1 \\ #2 \\ #3 \end{bmatrix}}
\newcommand{\techReport}[1]{(#1)}
\def\BibTeX{{\rm B\kern-.05em{\sc i\kern-.025em b}\kern-.08em
    T\kern-.1667em\lower.7ex\hbox{E}\kern-.125emX}}
\def \st { \mathbb{:}  }

\begin{document}

\title{\LARGE \bf Technical Report: A New Hopping
Controller for Highly Dynamical Bipeds*

\author{Shane Rozen-Levy$^{1}$  and Daniel E. Koditschek$^{2}$ 
}

\thanks{*This work was supported in part by the US Army Research Office under grant W911NF-17-1-0229.}
\thanks{$^{1}$Mechanical Engineering and Applied Mechanics, University of Pennsylvania, PA, USA. {\tt\small srozen01@seas.upenn.edu}}%
\thanks{$^{2}$Electrical and Systems Engineering, University of Pennsylvania, PA,USA.
        {\tt\small kod@seas.upenn.edu}}
}

\maketitle
\thispagestyle{empty}
\pagestyle{empty}

\begin{abstract}
We present angle of attack control, a novel control strategy for a hip energized Penn Jerboa. The energetic losses from damping are counteracted by aligning most of the velocity at touchdown in the radial direction and the fore-aft velocity is controlled by using the hip torque to control to a target angular momentum. The control strategy results in highly asymmetric leg angle trajectories, thus avoiding the traction issues that plague hip actuated SLIP. Using a series of assumptions we find an analytical expression for the fixed points of an approximation to the hopping return map relating the design parameters to steady state gait performance. The hardware robot demonstrates stable locomotion with speeds ranging from 0.4 m/s to 2.5 m/s (2 leg lengths/s to 12.5 leg lengths/s) and heights ranging from 0.21 m to 0.27 m (1.05 leg lengths to 1.35 leg lengths). The performance of the empirical trials is well approximated by the  analytical predictions. 
\end{abstract}

\section{Introduction}

\begin{figure}
     \centering
     \begin{subfigure}[t]{0.48\columnwidth}
         \centering
         \includegraphics[width=\columnwidth]{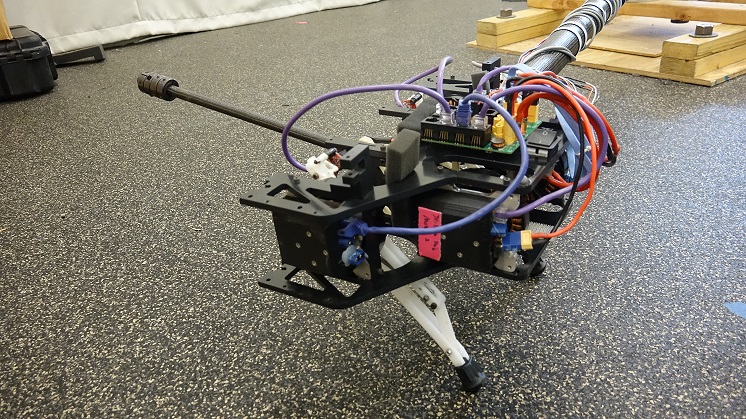}
         \caption{The Penn Jerboa \cite{shamsah_analytically-guided_2018}, a tailed biped with springy legs and only four actuators: one driving the leg angle at each hip; and two controlling the 2 DoF tail.}
         \label{TR-fig:jerboa}
     \end{subfigure}
     \hfill
     \begin{subfigure}[t]{0.48\columnwidth}
         \centering
         \includegraphics[width=\columnwidth]{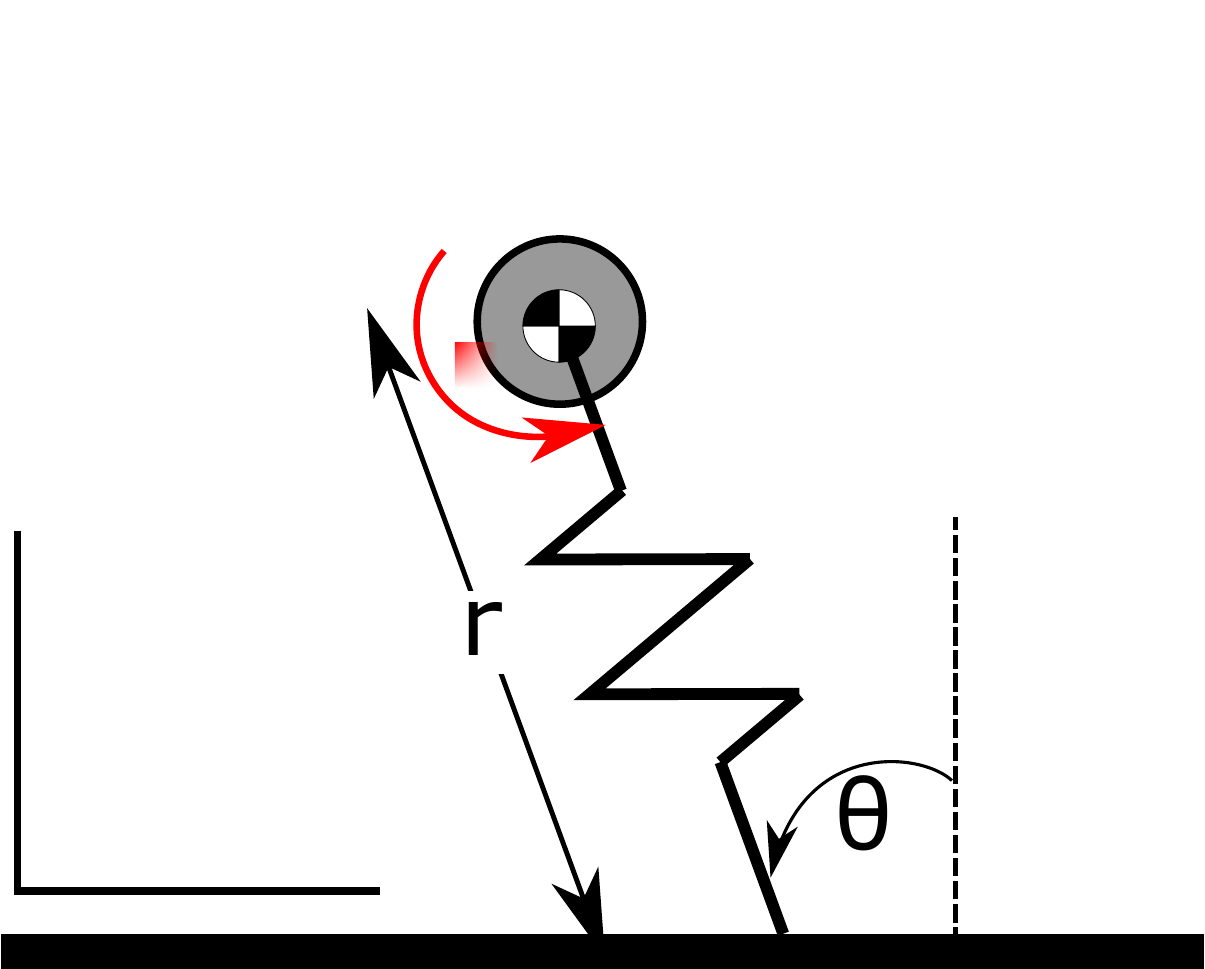}
         \caption{SLIP in stance. The black variables are states, the blue variables are parameters, and the red variables are control inputs.}
         \label{TR-slip}
     \end{subfigure}
        \caption{The Penn Jerboa and a schematic of the SLIP model used to analyze its 2 DoF sagittal plane hopping dynamics when fixed to a circular boom with locked pitch and an immobilized tail.  }
        \label{TR-fig:robot models}
\end{figure}

In contrast to many popular contemporary legged robots \cite{bledt_mit_2018, katz_mini_2019,hutter_anymal_2016}, the Penn Jerboa (fig. \ref{TR-fig:jerboa}) is a dramatically underactuated biped: it has 12 DoF (degrees of freedom) and only 4 direct drive actuators \cite{de_penn_2015, de_parallel_2015, kenneally_design_2016}. Moreover,  unlike many bipeds,
it features small point toes rather than flat feet  \cite{apgar_fast_2018}.  Jerboa's consequent high power density (43.2 w/kg \cite{kenneally_design_2016}) and its unusual recourse to a high powered (two DoF) tail at the expense of  affording only one actuator at each hip of its passive spring loaded legs provokes the question of whether and how its largely dynamical and comparatively more energetic regime of operation can offer performance competitive with that of more conventional legged designs.

Because its extreme underactuation precludes recourse to popular control approaches that assume the availability of arbitrary ground reaction forces subject to linear frictional constraints  \cite{samy_post-impact_2017, dai_planning_2016, bledt_implementing_2019, kim_highly_2019, carlo_dynamic_2018},  prior work on Jerboa has built on the long tradition of anchoring compositions of dynamical templates \cite{full_templates_1999, de_vertical_2018, kurtz_approximate_2020, kurtz_formal_2019}. This paper introduces a new entry to the catalogue of spring loaded inverted pendulum (SLIP) \cite{Schwind_Koditschek_1995} template controllers, offering a novel, model-relaxed strategy  for controlling the speed and height of Jerboa using its hip motors. This leaves the tail free to control the pitch and roll (future work). Compared to previous work on Jerboa, this control strategy allows the robot to hop faster with just as much height --- albeit potentially with the need for greater traction since the original tail-energized hopping mode \cite{shamsah_analytically-guided_2018} explicitly drives energy down the leg shaft, increasing the normal component of the ground reaction forces.

We find fixed points and prove stability for an approximate model of closed loop hopping generated by this control strategy by using assumptions to construct an analytical return map. The fixed points of this simplified system  effectively approximate the numerical fixed points of the unsimplified system allowing for an intuitive understanding of how the parameters affect the steady state operating regime of the robot. We test this control strategy in hardware showing a wide range of steady state operating regimes.

\subsection{Background Literature }
Since Raibert, the SLIP model has been used to study legged locomotion\cite{schwind_approximating_2000, seipel_simple_2007, altendorfer_stability_2004,saranli_rhex_2001, raibert_legged_1986}. 
Raibert controlled SLIP-like robots with a fixed thrust in stance to energize hopping and used the touchdown angle to control speed.

 Due to the simple nature of the flight map and resets, much of the analytical work on SLIP has focused on approximating the stance map \cite{geyer_spring-mass_2005, ankarali_analytical_2009}. The fixed points of the analytical return map serve as a map from the many dimensional parameter space to the operating conditions of the robot \cite{Koditschek_Buehler_1991}.

Inspired by the work in \cite{geyer_spring-mass_2005}, Ankarali et al. develops an analytical approximation of the stance map for SLIP with damping and gravity by linearizing gravity, yielding a conservation of angular momentum \cite{ankarali_analytical_2009}. Ankarali et al. then approximates the effect of gravity by iteratively estimating the average angular momentum \cite{arslan_approximate_2009}.

They extend the results to the stance map for hip energized SLIP  by taking the open loop hip torque into account when calculating the average angular momentum\cite{ankarali_stride--stride_2010,ankarali_analysis_nodate}. They use inverse dynamics to find the hip torque to get to the target energy and solve an optimization problem for the touchdown angle. Ankarali et al. are unable to analytically find fixed points of the return map and they never implemented the control strategy on a physical robot.

In addition to the work in \cite{ankarali_stride--stride_2010,ankarali_analysis_nodate}, there have been a handful of papers exploring hip energized hopping for SLIP-like robots\cite{cherouvim_control_2009, vasilopoulos_compliant_2014, shen_fundamental_2012}.

\subsection{Contributions and Organization}
With the goal of developing a simple, analytically tractable, control strategy for a hip energized SLIP-like robot, this paper makes three contributions: (i) a simple and novel control strategy for hip energized SLIP where energetic losses from damping are counteracted by choosing a touchdown angle that puts most of the speed at touchdown in the radial direction (eq. \ref{TR-eq:aoa}) and the leg angle is energized using the hip torque (eq. \ref{TR-momcon}); (ii) a closed form analytical approximation of the fixed points for a simplified model of SLIP under the new control strategy (eq. \ref{TR-eq:fixed}); (iii) an extensive empirical study of the implementation of the novel control strategy on the physical Jerboa robot \cite{shamsah_analytically-guided_2018}. 

This control strategy (i) allows for intuitive control of the fore-aft velocity(figure \ref{TR-fig:aoa_speed}) and some control over the apex height (figure \ref{TR-fig:aoa_height}). The analytical predictions (ii) match the fixed points of the numerical return map(table \ref{TR-tab:fixed_error_large}) and predict with reasonable accuracy the height and speed of the  robot (fig \ref{TR-fig:empricial_speed}, fig \ref{TR-fig:empirical_height}, table \ref{TR-tab:fixed_error_sim}). The robot (iii) demonstrated stable locomotion with speeds ranging from 0.4 m/s to 2.5 m/s (2 leg lengths/s to 12.5 leg lengths/s) and heights ranging from 0.21 m to 0.27 m (fig \ref{TR-fig:empricial_speed}, fig \ref{TR-fig:empirical_height}, fig \ref{TR-fig:steady_state}).

Section \ref{TR-sec: return} develops the approximations that yield a closed form return map. Section \ref{TR-sec:fixed} introduces further simplifying assumptions affording simple closed form approximation of the fixed points of the  return map and compares the accuracy of both relative to their counterparts arising from numerical integration of the original physical model (sec. \ref{TR-sec:accuracy_fixed}). Section \ref{TR-sec:sim} demonstrates the performance of the controller on Jerboa.  Section \ref{TR-sec:conclusion} wraps up the paper with a discussion of the key ideas and some future steps.

\begin{table}[ht]
\small\sf\centering
\begin{tabular}{lll}
\toprule
Symbol&Brief Description & Ref\\ 
\midrule
$\theta$& leg angle & fig. \ref{TR-slip} \\
$r$& leg length & fig. \ref{TR-slip}\\
$m$ & mass  &  fig. \ref{TR-slip}\\
$k$&  spring constant  & fig. \ref{TR-slip} \\
$\tau$& hip torque  & fig. \ref{TR-slip} \\
$b$&  damping coefficient  & fig. \ref{TR-slip} \\
$r_0$& spring rest length &  fig. \ref{TR-slip} \\
$x$ & fore-aft position & fig. \ref{TR-slip} \\
$y$ & height relative to ground  & fig. \ref{TR-slip} \\ 

$z^s$ & the state in stance = $[r, \dot{r}, \theta, \dot{\theta}]$ & sec. \ref{TR-stance_map} \\

$g$ & acceleration due to gravity & eq. \ref{TR-slip_dyn}\\\

$k_\theta$ & angle of attack gain &  eq. \ref{TR-eq:aoa}\\
$\theta_\text{AoA}$ &angle of attack & eq. \ref{TR-eq:aoa}\\

$\bar{p}_\theta$ & target angular momentum in stance &  eq.  \ref{TR-momcon}\\

$z^a$ & the state at apex = $[\dot{x}, y]$ & eq. \ref{TR-eq:return}\\
\bottomrule
\end{tabular}
\caption{Symbol definitions\label{TR-T2}}

\end{table}

\section{Return Map}
\label{TR-sec: return}
 The ballistic modes (ascent and descent) of the SLIP model are completely integrable; hence the effort in constructing the return map whose fixed points are of interest lies in approximating its non-integrable stance mode map  \cite{Ghigliazza_Altendorfer_Holmes_Koditschek_2003}.

\subsection{Stance Map, Ascent Map, Descent Map, Resets}
\label{TR-stance_map}

The non-integrability \cite{holmes_poincare_1990} of SLIP dynamics
\begin{align}
    \begin{bmatrix}
    \ddot{r} \\
    \ddot{\theta}
    \end{bmatrix} = \begin{bmatrix}
     - k/m (r-r_0)-b/m \dot{r} -g \cos(\theta)  \\
     \frac{-2 \dot{r} \dot{\theta}}{r} + g/r \sin (\theta) + \frac{\tau}{m r^2}
    \end{bmatrix} \label{TR-slip_dyn}
\end{align}
introduces the burden of  imposing simplifying assumptions   yielding physically effective approximations for  designers seeking closed form expression and stability guarantees for steady state gaits.

In order to get integrable dynamics we start with two assumptions inspired by the work in \cite{arslan_approximate_2009}.

\begin{assumption}
Since we control to a target angular momentum, $\bar{p_\theta}$, the angular momentum is constant and equal to the target angular momentum. \label{TR-ass_mom}
\end{assumption}

\begin{assumption}
The leg angle is small, thus gravity acts radially. \label{TR-ass_grav}
\end{assumption}

 As in \cite{arslan_approximate_2009}, we take a Taylor series approximation of the $1/r^3$ and $1/r^2$ terms in the post assumption dynamics \techReport{eq. \ref{TR-post_assum_dyn}}  centered at $r=r_g$, where $r_g = r0-mg/k$, yielding

\begin{align}
    \begin{bmatrix}
    \ddot{r} \\
    \dot{\theta}
    \end{bmatrix} = \begin{bmatrix}
     \frac{\bar{p_\theta}^2}{m^2r_g^3}-(\frac{3\bar{p_\theta}^2}{m^2r_g^4}+\frac{k}{m})(r-r_g)-\frac{b}{m}\dot{r}  \\
     \frac{3\bar{p_\theta}}{m r_g^2} - \frac{2\bar{p_\theta}}{mr_g^3}r 
    \end{bmatrix}, \label{TR-taylor}
\end{align}
a  1 DoF linear time invariant damped harmonic oscillator in $r$ feeding forward to excite a first order linear time invariant leg angle integrator in $\theta$. 
 For the closed form solutions see \techReport{eq. \ref{TR-r_final}, eq. \ref{TR-dr_final}, eq. \ref{TR-theta_final}}.

Integrating (\ref{TR-taylor}) and following \cite{ankarali_stride--stride_2010} to obtain  $t_{lo}$, the liftoff time, yields the stance map approximation, $\Psi_s \st \mathbb{R}^3 \times S^1 \rightarrow \mathbb{R}^3 \times S^1$,  $ \st  z^s_{td} \mapsto z^s_{lo}$, which we display in closed form below in eq. \ref{TR-simple_stance} after introducing a number of further simplifying assumptions. 
\subsubsection{Ascent, Descent, and Reset Maps}
\label{TR-sec:flight}
Due to the single point mass, SLIP follows a ballistic trajectory in flight. Apex is defined by $\dot{y} = 0$, and touchdown occurs when the toe contacts the ground.

 Similarly due to the massless foot and springy legs, the reset maps do not have to handle impacts. Thus, the resets convert between the polar coordinates used in stance and the cartesian coordinates used in flight. See  {\techReport{app. \ref{TR-appendix:flight}, app. \ref{TR-appendix:resest}}} for details on the resets, ascent, and descent maps.

\subsection{Angular and Radial Control Policies}
 Early on in our experiments, we found that Raibert stepping \cite{raibert_legged_1986} did not work well for hip energized hopping because the Coriolis term did not provide sufficient coupling for the hip torque to counteract the radial damping.

Angle of attack control  (hereafter, AoA) counteracts the energetic losses from damping by choosing a touchdown angle that puts most of kinetic energy at touchdown into the $\dot{r}$ component; as a result $|\dot{\theta}_{td}|$ is small. The hip torque then re-energizes $\theta$ directly, without relying on coupling. 

Let $\theta_{td} = k_\theta \theta_\text{AoA}$ be the touchdown angle under AoA where $k_\theta$ is the gain on the touchdown angle (nominally 1) and $\theta_\text{AoA}$ is the angle of attack. $\theta_\text{AoA} := \arctan{\frac{\dot{x}_{td}}{\dot{y}_{td}}}$.
Since $\dot{y}_{td}$ varies with $\theta_{td}$, this yields the constraint equation \begin{equation}
     \theta_\text{AoA} =\Phi(\theta_\text{AoA}) :=\arctan\frac{\dot{x}}{\sqrt{2 E_v/m - 2 g r_0 \cos(k_\theta \theta_\text{AoA})}}, \label{TR-eq:aoa}
\end{equation}
 where $E_v := 1/2 m \dot{y}^2 + m g y$ is the vertical energy. An approximation to the implicit function for $\theta_{AoA}$ satisfying constraint \ref{TR-eq:aoa} is presented  in \techReport{eq. \ref{TR-Appendix:aoa_approx}}.
 
The AoA gain, $k_\theta$, is nominally between $0$ and $1$. $k_\theta < 1$ corresponds to having some initial angular velocity at touchdown. A $k_\theta > 1$ corresponds to having angular velocity in the opposite direction of travel at touchdown. For this paper we will restrict $k_\theta \in [0,1]$.

The angular control policy is a PID + feed forward loop which control the leg to a target angular momentum.
\begin{equation}
    \tau_j = k_p (\Bar{p}_\theta - p_{\theta,j}) + k_i \sum_{i=0}^j (\Bar{p}_\theta - p_{\theta,i})  - k_d \dot{p}_{\theta,j}  - m g r_j \sin{\theta_j} \label{TR-momcon}
\end{equation}

\subsection{Constructing the Return Map}
Given the stance map ($\Psi_s$, sec \ref{TR-stance_map}), resets(${^sR^f}, {^fR^s}$, sec. \ref{TR-sec:flight}), ascent map($\Psi_a$, sec. \ref{TR-sec:flight}), and descent map($\Psi_d$, sec. \ref{TR-sec:flight}), we construct the apex coordinate return map by composing the analytical functions. 
\begin{align}
    P:& \mathbb{R}^2 \rightarrow \mathbb{R}^2 := \Psi_a \circ {^fR^s} \circ \Psi_s \circ {^sR^f} \circ \Psi_d  \nonumber \\
    \st &z^a_k \mapsto z^a_{k+1} \label{TR-eq:return}
\end{align}

\section{Fixed Points and Stability of the Analytical Return Map}
\label{TR-sec:fixed}
Notwithstanding the closed form expression for $P$ represented by (\ref{TR-eq:return}), developing an intuitively useful closed form expression for its fixed points is facilitated by the following simplifying assumptions. 

\begin{assumption}
Across all fixed points, the nondimensionalized time of stance, $t_{lo}/\omega_d$ is constant. \label{TR-ass_4}
\end{assumption}

\begin{assumption}
  $r_{lo} \approx r_0$. \footnote{Based on the radial velocity at liftoff, the spring constant, and the damping, this is accurate within 2mm.  } \label{TR-ass_6}
\end{assumption}

Using these assumptions the stance map, $\Psi_s$ derived in Sec. \ref{TR-stance_map}  takes takes the form
\begin{align}
    z^s_{lo} = 
    \begin{bmatrix} r_0 \\ C_1 \dot{r}_{td} + C_2 \\  \theta_{td} + C_3 \dot{r}_{td} + C_4 \\ \bar{p}_\theta/(m r_0^2)
    \end{bmatrix}, \label{TR-simple_stance}
\end{align}
where $C_1, ... C_4$ are defined in \techReport{eq. \ref{TR-C1}-eq. \ref{TR-C4}}.

\begin{assumption}
 $\cos{\theta} \approx 1-\frac{\theta^2}{2}$ and $\sin{\theta} \approx \theta$ (see figure \ref{TR-fig:steady_state} for evidence that this assumption is valid ). \label{TR-ass_5}
\end{assumption}

\begin{assumption}
The change in gravitational potential energy between touchdown and liftoff is negligible compared to the kinetic energy
\label{TR-ass_energy}
\end{assumption}

\begin{assumption}
$\dot{\theta}_{td} \approx  -\dot{r}_{td}/r_0\tan{(\theta_\text{offset})}$ where $\theta_\text{offset} := \Bar{p}_\theta/0.7\frac{\pi}{4}(1-k_\theta)$. 
\label{TR-ass_aoa}
\end{assumption}

See {\techReport{app. \ref{TR-Appendix_aoa_assumption}}} for a derivation of this assumption. Total energy and fore-aft velocity are conserved in flight, thus
 at the fixed point we impose the constraints, $E_{td} = E_{lo}$ and $\dot{x}_{td} = \dot{x}_{lo}$ yielding quadratic polynomials in the touchdown state variables {\techReport{eq. \ref{TR-e1}, eq. \ref{TR-dx2}}}. The constraint on energy yields  a quadratic polynomial in  $\dot{r}^*_{td}$ whose coefficients, $a_r, b_r, c_r$ are given by elementary functions of the physical parameters and control inputs listed in \techReport{eq. \ref{TR-r_coeff}}. In turn, the constraint on fore-aft velocity yields a quadratic polynomial in $\theta^*_{td}$ whose coefficients, $a_\theta, b_\theta, c_\theta$ are given by similar functions that now also depend upon the value of $\dot{r}^*_{td}$ \techReport{eq. \ref{TR-a_theta}-eq. \ref{TR-b_theta}}. Thus, the fixed points of $P$ (eq. \ref{TR-eq:return}) are given as 

\def \drtd {  \dot{r}^*_{td}}
\def \thtd { \theta^*_{td} }
\def \dthtd {\dot{\theta}^*_{td}}
\def \args { \bar{p}_\theta, k_\theta}

\begin{align}
&\vthre{ \drtd }{ \thtd } { \dthtd } =F (\args) 
:= \vthre{R (\args) )}{\Theta( \args) }  {\Xi( \args) }
\end{align}
where
\begin{align*}
  R( \args)& := Q_{-} \big( a_r(\args), b_r, c_r(\bar{p}_\theta) \big) \\
 \Theta( \args)& := Q_{+} \big(  a_\theta ( R(\args), \args), \\ 
&b_\theta ( R(\args), \args), c_\theta(  R(\args), \args) \big) \\
  \Xi (\args) &:= - R(\args) / r_0 \tan(\theta_\text{offset})
\end{align*}
where $Q_\pm$ are the positive and negative roots of a quadratic polynomial \techReport{eq. \ref{TR-quadratic_root}}.

Given the fixed point in touchdown coordinates, we map it backwards in time to get the apex coordinate fixed points:
\begin{equation}
    {z^a}^* = \Psi_d^{-1} \circ ({^sR^f})^{-1} ([r_0, F( \args)]) \label{TR-eq:fixed}
\end{equation}

We check stability of the fixed points by evaluating the Jacobian of the return map and checking if its spectral radius is less than 1 at the fixed point.

\subsection{Plots of Analytical Fixed Points}

\begin{figure*}
\centering
\begin{minipage}[t]{\columnwidth}
    \includegraphics[width=\columnwidth]{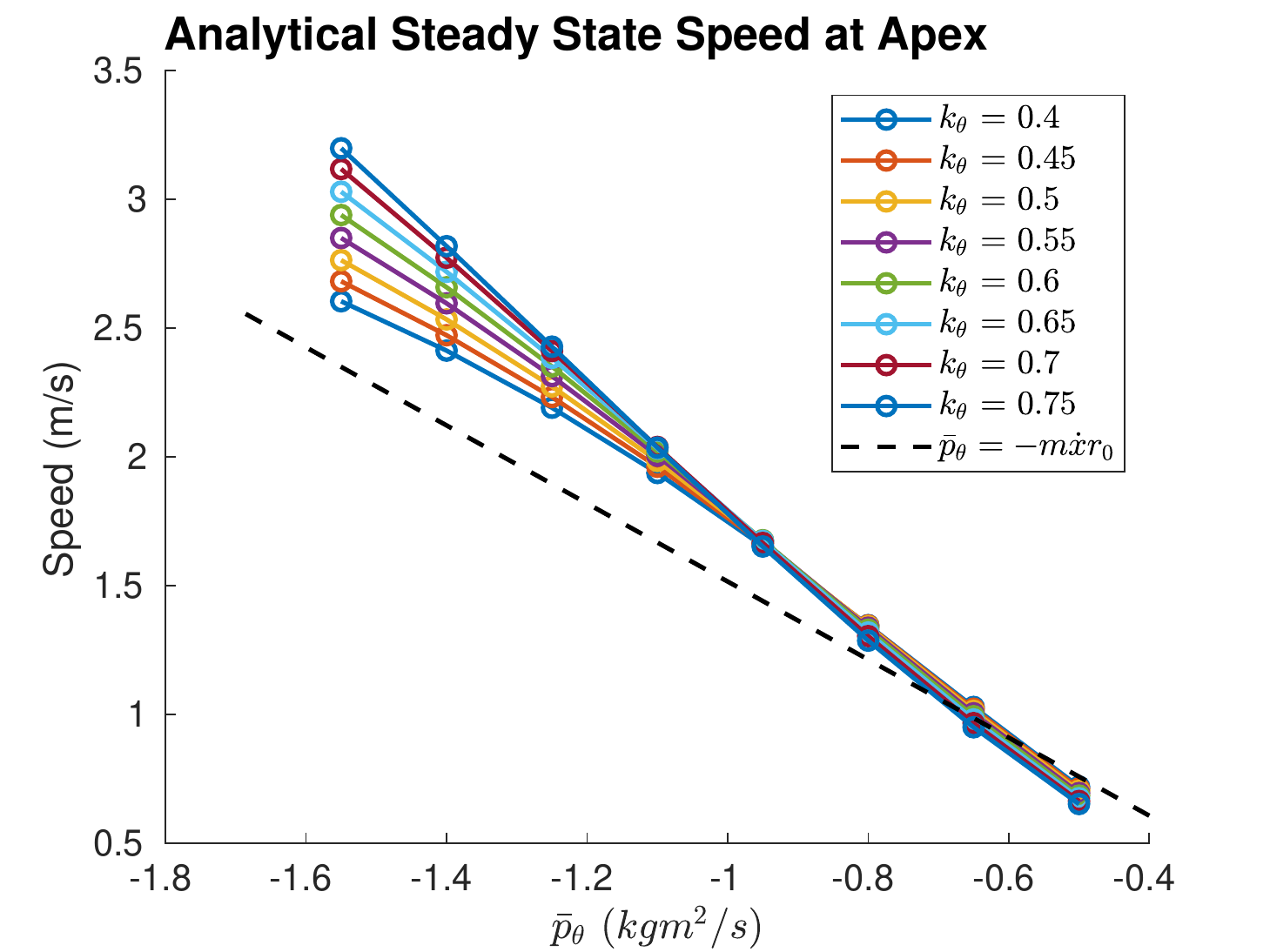}
\caption{Apex velocity of fixed points predicted by the analytical return map (eq. \ref{TR-eq:fixed}) as functions of the control inputs $k_\theta$ and $\Bar{p}_\theta$. This plot shows a decoupling between $k_\theta$ and $\Bar{p}_\theta$; increasing $|\Bar{p}_\theta|$ increases the speed while changing $k_\theta$ has almost no effect on the speed.  $\Bar{p}_\theta = -m \dot{x}_\text{des} r_0$ is the line corresponding to the fore-aft velocity if $\theta_{lo} =0$. The accuracy of these fixed points relative to the numerically integrated values is presented in table \ref{TR-tab:fixed_error_large}. The model parameters are in table \ref{TR-parameters}. }
\label{TR-fig:aoa_speed}
\end{minipage}\hfill
\begin{minipage}[t]{\columnwidth}
    \includegraphics[width=\columnwidth]{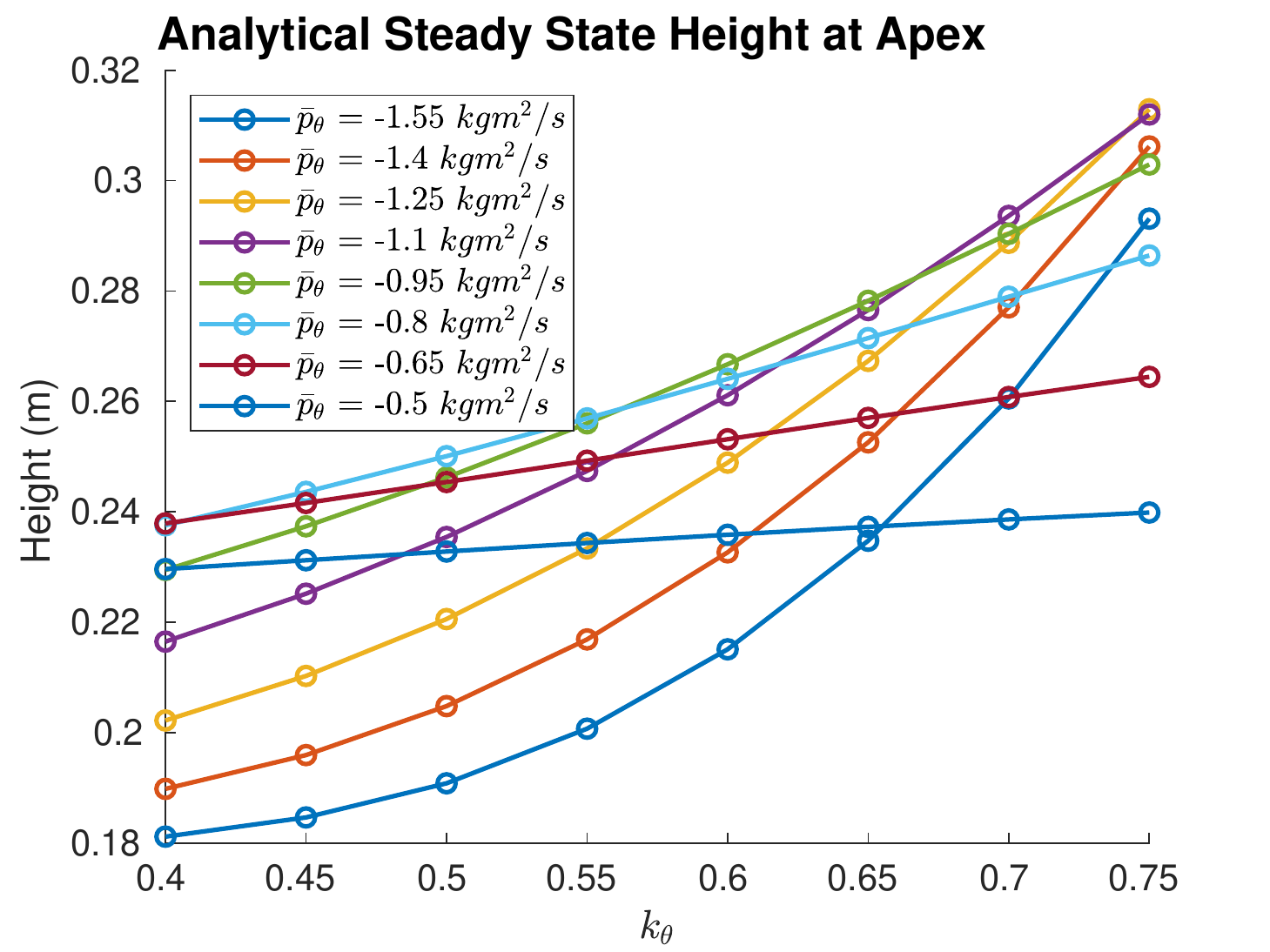}
\caption{ Apex height of fixed points predicted by the analytical return map  (eq. \ref{TR-eq:fixed}) as a function of the control inputs $k_\theta$ and $\Bar{p}_\theta$. Increasing $k_\theta$ increases the apex height.  Contrary to $k_\theta$, $\bar{p}_\theta$'s effect on the height is not monotonic. For a fixed $k_\theta$, there is a $\bar{p}_\theta$ resulting in a maximum height.  This corresponds to $\theta_{lo} = 0$, thus a higher $|\bar{p}_\theta|$ decreases $\dot{y}_{lo}$ and a lower $|\bar{p}_{\theta}|$ results in less energy to put into the radial subsystem.  The accuracy of these fixed points relative to the numerically integrated values is presented in table \ref{TR-tab:fixed_error_large}. The model parameters are in table \ref{TR-parameters}.   }
\label{TR-fig:aoa_height}
\end{minipage}
\end{figure*}

Table \ref{TR-parameters} has the parameters used in the analytical return map. Figure \ref{TR-fig:aoa_speed} and \ref{TR-fig:aoa_height} plot the apex fore-aft speed and height as a function of the control inputs, $\Bar{p}_\theta$ and $k_\theta$. The apex speed shows a significantly decoupling between $k_\theta$ and $\Bar{p}_\theta$; $\Bar{p}_\theta$ heavily effects the speed while $k_\theta$ has almost no effect on the speed. By using $\Bar{p}_\theta = -m \dot{x}_\text{des} r_0$ (corresponding to speed when $\theta_{lo} = 0$), the resulting relationship between the target and actual speed would be monotonic and zero at zero target speed.

 \begin{table}[ht] \centering
 \begin{tabular}{|l|l|}
\hline
$m$ (kg)                      & $3.3$\\ \hline
$r_0$ (m)                      & 0.2  \\ \hline
$k$ (N/m)                     & 4000 \\ \hline
$b$ (Ns/m)                    & 20   \\ \hline
\end{tabular}
\caption{Jerboa's model parameters}
\label{TR-parameters}
\end{table}

On the other hand, the apex height shows some coupling between $k_\theta$ and $\Bar{p}_\theta$. $k_\theta$ has a monotonic relationship with height, though increasing $|\bar{p}_\theta|$ increases $\partial y^* / \partial k_{\theta}$. This should be thought of as for the higher $|\bar{p}_\theta|$, $k_\theta$ has a larger affordance on height. The simple formulation of this control strategy, the decoupling in the control of fore-aft velocity, and the angle of attack gain's consistent effect on height makes this control strategy ideal for hip energized SLIP.

\subsection{Accuracy of the Fixed Points}
\label{TR-sec:accuracy_fixed}
In order to validate our numerous assumption and  evaluate the accuracy of the analytical return map, we compared its fixed points (eq. \ref{TR-eq:fixed}) to the fixed points of the numerical return map.

Table \ref{TR-tab:fixed_error_large} is the error between the fixed points of the analytical return map and the fixed points of the numerical return map over a rectangular approximation of the robot's operating regime,  $\Bar{p}_\theta \in [-0.5, -1.55] kg m^2/s$, $k_\theta \in [0.4, 0.75]$. The error is mostly small over the operating regime except for the error in $\dot{x}$ when $|\bar{p}_\theta|$ and $k_\theta$ are large. These control inputs result in a large $\theta_{td}$, thus violating assumptions \ref{TR-ass_grav}, \ref{TR-ass_5},  and \ref{TR-ass_energy}.

\begin{table}\centering 
\begin{tabular}{|l|l|l|}
\hline
       & RMS Error   & Percent RMS Error  \\ \hline
$\dot{x}$      & 0.488 m/s & 20.3\%           \\ \hline
$y$     & 0.037 m   & 13.3\%            \\ \hline
\end{tabular}
\caption{Accuracy of the analytical return map's predicted speed and height (eq. \ref{TR-eq:fixed}) compared to the numerical return map's predicted speed and height over a rectangular approximation of the robot's operating regime. $\Bar{p}_\theta \in [-0.5, -1.55] kg m^2/s$, $k_\theta \in [0.3, 0.75]$. The RMS error in the apex coordinates is small showing that the approximate analytical return map matches the  numerical return map. The model parameters are in table \ref{TR-parameters}. See \techReport{fig \ref{TR-fig:fixedPointLargeError}} for plot of the error of the analytical return map's fixed points compared to the numerical return map's fixed points.}
\label{TR-tab:fixed_error_large}
\end{table}

\section{Experiments}
\label{TR-sec:sim}
In order to test  angle of attack control, we implemented the controller on the  boom-mounted, pitch-locked, Jerboa robot with its tail removed  \cite{shamsah_analytically-guided_2018}. 

The control strategy works very well in hardware. The robot achieved speeds of up to 2.5 m/s (limited by kinematics) and heights of up to 0.27 m. The main failure modes were premature touchdown and failing to liftoff. The attached supplemental video has clips of Jerboa hopping at representative steady state set points across the operating regime. Fig. \ref{TR-fig:empricial_speed} and Fig. \ref{TR-fig:empirical_height} contrast the analytical prediction of apex coordinate fixed point with empirical data taken at steady state.

\subsection{Experimental Setup}
  The Jerboa robot weighs 3.3 kg, has a max hip torque of 7 Nm from 2 TMotor U8 \cite{tmotor}, a 4000 N/m spring leg, and an operating voltage of 16.8 V from an offboard LiPo. 
 Jerboa's processor is a PWM mainboard from Ghost Robotics\cite{noauthor_revolutionizing_nodate} which runs angle of attack control at 1 kHz.

\subsubsection{Implementation Detail}
Angle attack control requires an estimate of the robot's velocity at liftoff (\ref{TR-eq:aoa}). We estimated this value using a coarse derivative of the robot's position estimated with the leg kinematics (touchdown to liftoff for fore-aft velocity, and bottom to liftoff for vertical velocity).

Additionally, the motors controllers on Jerboa did not have current/torque control, only voltage control. Fortunately we were still able to use a controller of the same form as eq. $\ref{TR-momcon}$, though the constant on the gravity compensation term had to be manually tuned.

\subsection{Empirical Fixed Points}
\begin{figure*}
\centering
\begin{minipage}[t]{\columnwidth}
    \includegraphics[width=\columnwidth]{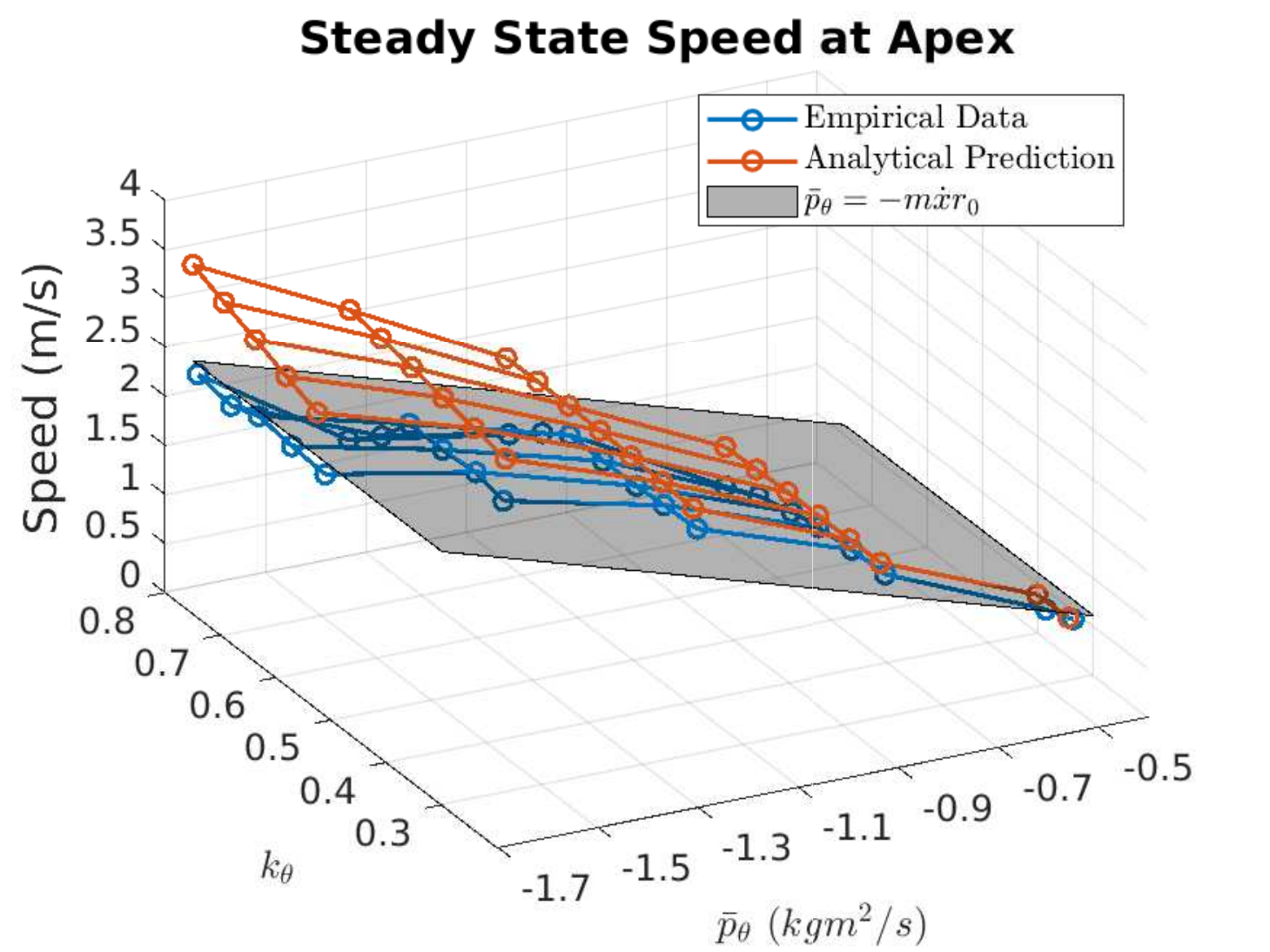}
\caption{The empirical speed of Jerboa compared to the speed predicted by the analytical return map (eq. \ref{TR-eq:fixed}) over the robot's operating regime. Increasing $|\bar{p}_\theta|$ increases the speed in both the empirical data and the analytical predictions. Due to the assumptions breaking down at the higher energy levels and the robot hitting its kinematic limits, the analytical return map over estimates the velocity at the higher target speeds.}\label{TR-fig:empricial_speed}
\end{minipage}\hfill
\begin{minipage}[t]{\columnwidth}
    \includegraphics[width=\columnwidth]{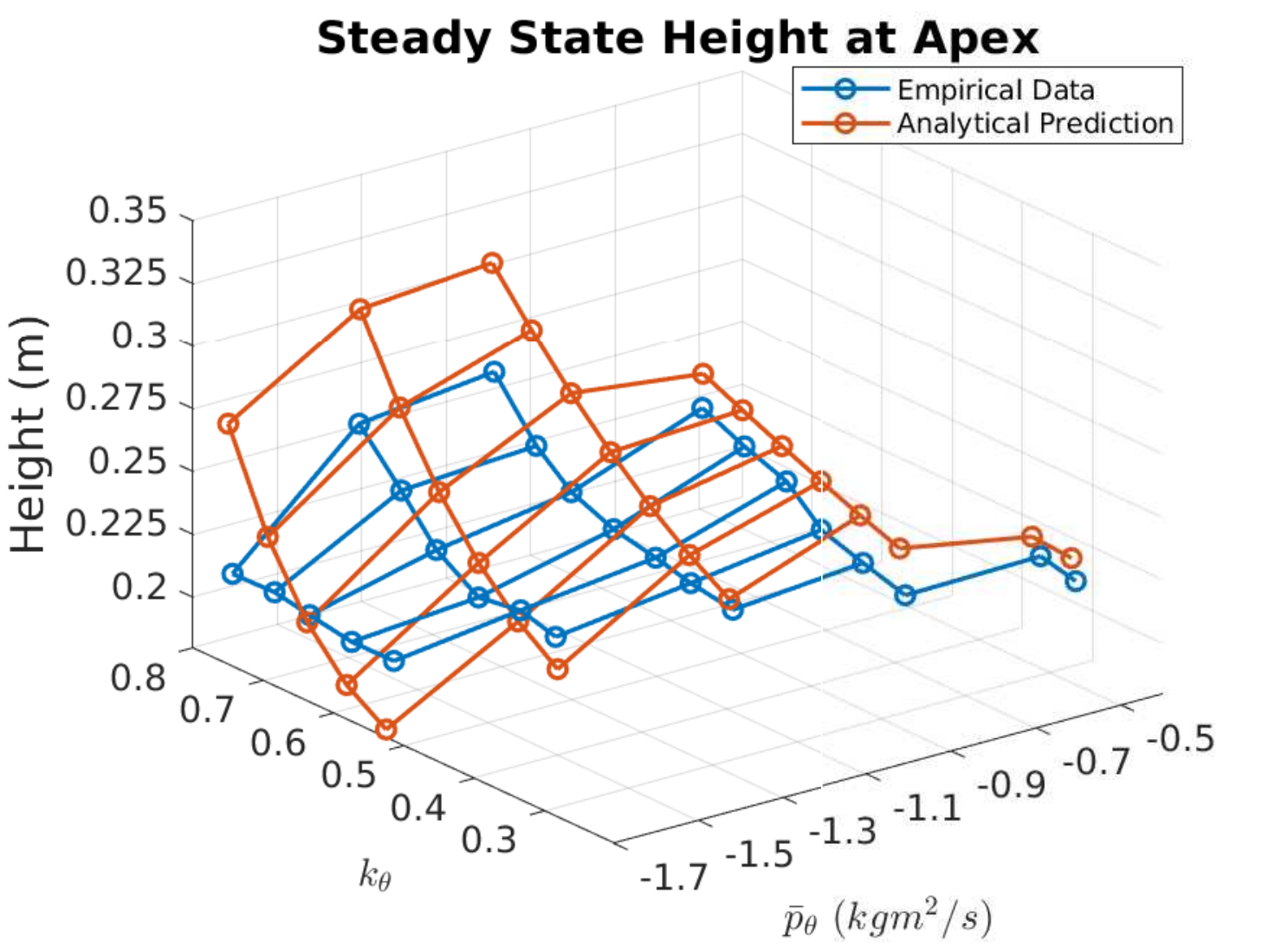}
\caption{The empirical height of Jerboa compared to the height predicted by the analytical return map (eq. \ref{TR-eq:fixed}) over the robot's operating regime. Many of the trends present in the analytical predictions are present in the empirical data. For example as $k_\theta$ increases, the height also increases.
Due to the assumptions breaking down at the higher energy levels and the robot hitting its kinematic limits, the analytical return map over estimates the height at the higher $|\bar{p}_\theta|$ and $k_\theta$. }\label{TR-fig:empirical_height}
\end{minipage}
\end{figure*}

We tested the robot with $\bar{p}_\theta \in [-0.4, -1.7] kg m^2/s$ and $k_\theta \in [0.3, 0.75]$. Figure \ref{TR-fig:empricial_speed} and \ref{TR-fig:empirical_height} plot the apex coordinate fixed points for the control inputs that resulted in stable locomotion. Jerboa was able to hop at speeds ranging from 0.4 m/s (2 leg lengths/s) to 2.5 m/s (12.5 leg length/s). As was predicted by the analytical return map, the fore-aft speed is controlled by $\bar{p}_\theta$. Additionally, the robot was able to hop at heights ranging from 0.21 m to 0.27 m. As with the analytical return map, the height is controlled by the $\bar{p}_\theta$ and $k_\theta$ where increasing $k_\theta$ increases the height. Table \ref{TR-tab:fixed_error_sim} presents the error of the fixed points of the analytical return map compared to the empirical results.

    \begin{table}\centering 
\begin{tabular}{|l|l|l|l|l|}
\hline
       & RMS Error   & Percent  & Max Error & Max Percent   \\
       &             &RMS Error &           & Error \\\hline
$\dot{x}$      & 0.595 m/s & 36.7\%      & 1.328 m/s   & 82.0\%   \\ \hline
$y$     & 0.0264 m   & 11.6\%       & 0.060 m    & 26.5\%   \\ \hline
\end{tabular}
\caption{Accuracy of the analytical  apex coordinate fixed points compared to the empirical fixed points over the robots operating regime. The analytical return map does an excellent job predicting $\dot{x}$ and $y$ in the low and middle operating range and deteriorates at the higher ranges of $k_\theta$ and $\bar{p}_\theta$ as shown in Figs. \ref{TR-fig:empricial_speed} \& \ref{TR-fig:empirical_height}. }
\label{TR-tab:fixed_error_sim}
\end{table}

The analytical return map does an excellent job predicting the empirical fixed points. Not only are the trends preserved, but the actual values are fairly similar. The height is accurate to about 1/10 of a leg length and the error in speed is concentrated at the higher angular momentums.

\subsection{Steady State Trajectories}

\begin{figure}
    \centering
    \includegraphics[width=\columnwidth]{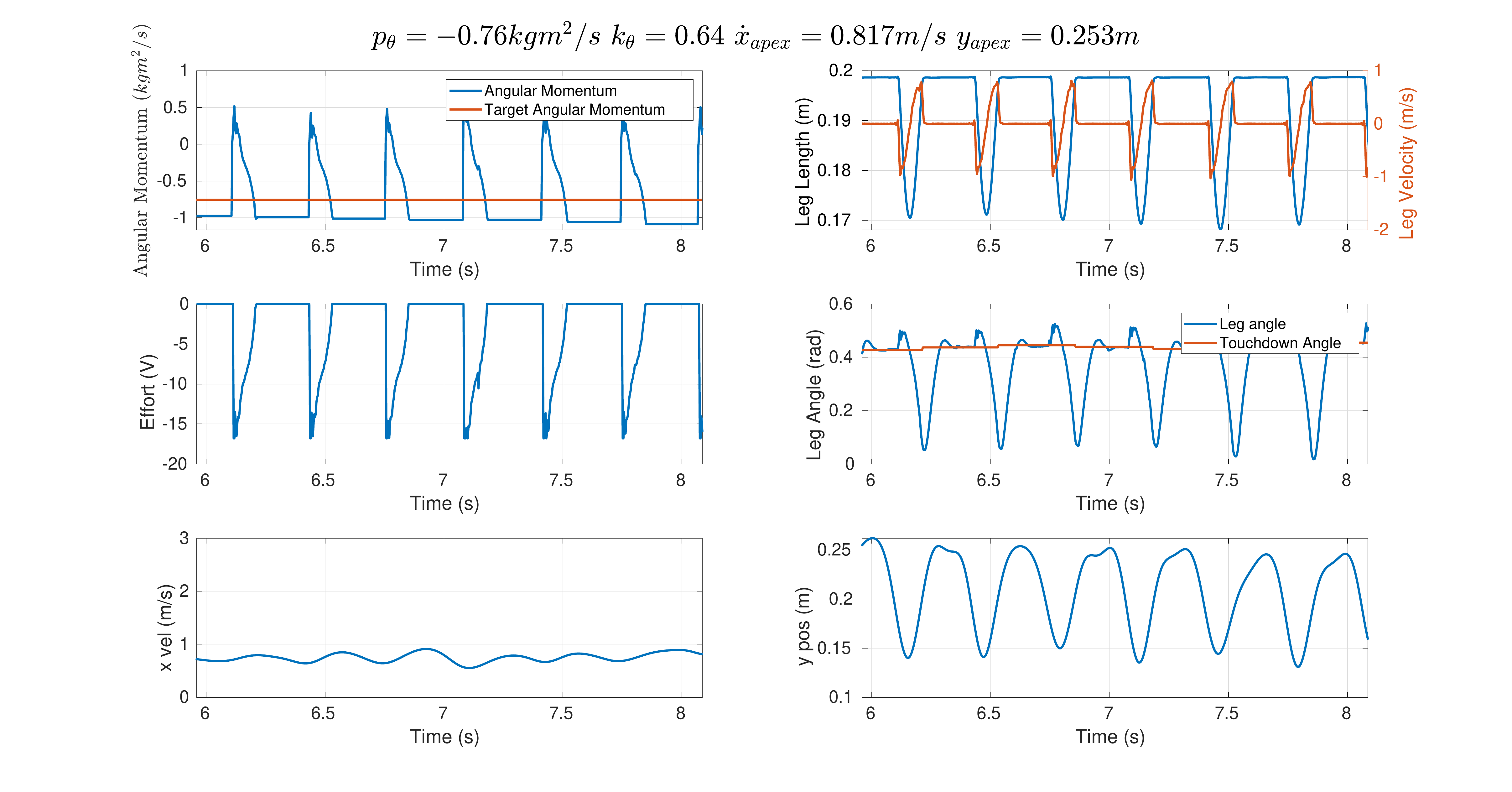}
    \caption{ A time domain trace of the robot hopping with $\bar{p}_\theta = -0.79 kgm^2/s$ and $k_\theta = 0.64$. At touchdown the hip motors briefly saturate before dropping down to a lower voltage. Additionally, the leg angle trajectory is highly asymmetric ($\theta_{td} = 0.45 rad, \, \theta_{lo} = 0.05 rad$) which prevents slippage.}
    \label{TR-fig:steady_state}
\end{figure}

Figure \ref{TR-fig:steady_state} shows the steady state trajectory of the robot with moderate speed and height.  The angular momentum asymptotically converges to the target angular momentum over the course of stance without saturating the motors. The resulting leg angle trajectories are very asymmetric. $\theta_{td} \approx 0.45 rad$ while $\theta_{lo} \approx 0 rad$. This means that the hip torque is always increasing the normal component of the ground reaction forces, rather than decreasing it. Even though $k_\theta < 1$,  ${p_{\theta}}_{td}$ is  opposite the direction of travel. This is likely caused by the impact at touchdown and the 4 bar linkage in the leg.

These plots also shows the validity of some of our earlier assumptions. For the lower $|\bar{p}_\theta|$, the leg angle stays small. Although we assumed angular momentum is constant, it changes quickly over stance, always reaching or slightly overshooting $\bar{p}_\theta$. 

\begin{figure}
    \centering
    \includegraphics[width=\columnwidth]{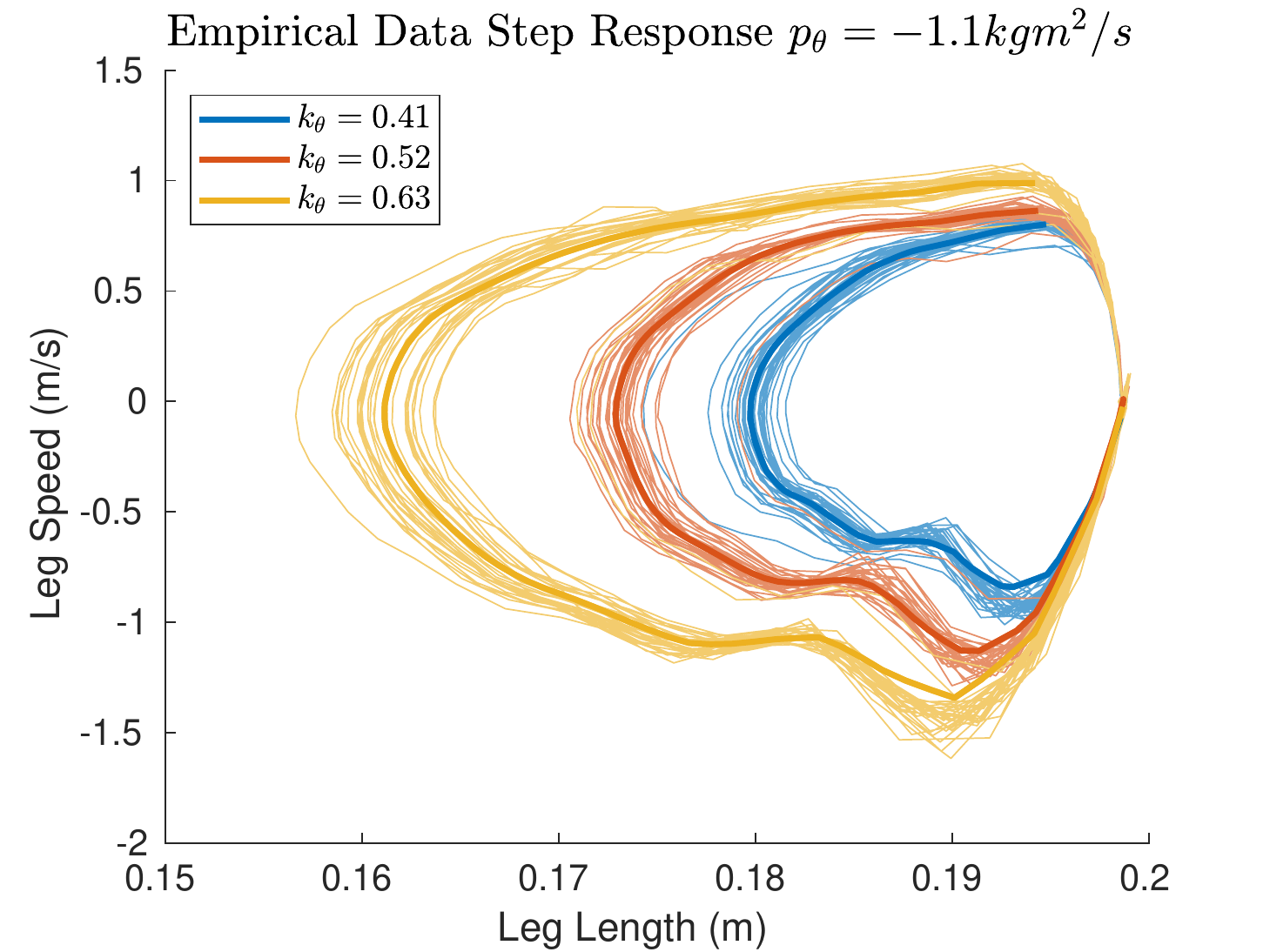}
    \caption{ The $r, \dot{r}$ limit cycle for $\bar{p}_\theta = -1.1 kgm^2/s$ and  a step response of $k_\theta$. As predicted by the analytical results increasing $k_\theta$ increases the energy in the radial subsystem.}
    \label{TR-fig:limitcycle}
\end{figure}

Figure \ref{TR-fig:limitcycle} is the $r, \dot{r}$ limit cycle from the empirical trials for $\bar{p}_\theta = -1.1 kgm^2/s$ and a step response of $k_\theta$. This plot shows that the system to robust to changes in input and that increasing $k_\theta$ increases the energy in the radial subsystem.

\section{Conclusion}
\label{TR-sec:conclusion}

This paper presents a novel hip energized control strategy for a pitch constrained Jerboa. The control strategy allows Jerboa to hop at speeds of up 2.5 m/s. By constructing and validating an analytical return map  we found a closed form expression for the fixed points giving insight into how the parameters affect the operating regime of the robot. 

\subsection{Discussion}

\label{TR-sec:discussion}
\subsubsection{Energizing the Radial Component Through Resets}
With hip energized SLIP, the energetic losses enter through the radial dynamics while the energization happens in the leg angle dynamics. Since the only coupling in stance occurs through the Coriolis terms of the dynamics (which are typically very weak in regimes of physical interest), we energize the radial direction through a smart choice of reset map.
By controlling around a touchdown angle that maximizes $|\dot{r}_{td}|$ (eq. \ref{TR-eq:aoa}), the radial direction can be energized at the cost of $\dot{\theta}_{td}$. The leg angle is then re-energized using a hip torque. One side effect of this strategy is that the asymmetry in the leg angle greatly reduces the traction concerns that comes with hip energized hopping \cite{de_modular_2017}.

\subsubsection{High Speed Hopping}
Angle of attack control allowed Jerboa to hop at speeds of 2.5 m/s, limited only by the maximum leg angle. As a point of comparison Research Rhex \cite{galloway_x-rhex_2010}, scaled to the size and mass of Jerboa, has a top speed of 2.35 m/s \cite{miller_dynamic_2015}.

\subsubsection{Comparison to Previous Jerboa Controller}
Compared to the previous tail energized Jerboa controllers \cite{de_penn_2015,de_modular_2017, shamsah_analytically-guided_2018,de_parallel_2015}, angle of attack control allows Jerboa to hop faster with just as much height. One downside of controlling a robot with angle of attack control is that it can't hop in place; thus when transitioning from forward hopping to backwards hopping the robot needs to transition from angle of attack control, to tail energized, and back. This maneuver points to: (i) situations where it makes sense to use one strategy vs. the other, and (ii) the need to develop ways of transitioning from tail energized hopping to angle of attack control.

\subsubsection{Model Relaxed Control}
Contrary to other hip energized SLIP controllers \cite{ankarali_stride--stride_2010,cherouvim_control_2009}, angle of attack control does not rely on any model parameters besides the mass and leg length, both of which are easy to measure. This makes the controller robust to changes in payload or the replacement of a broken leg. 

\subsection{Future Work}

Future work should focus on developing controllers for Jerboa that allow it to use angle of attack control without the constraints on pitch and roll. The previously developed roll controller \cite{wenger_frontal_2016} is a good place to start.

Additionally the high speed achieved on Jerboa makes angle of attack control a potentially good strategy for other robots. As a future work we propose implementing angle of attack control on robots such as X-RHex or Ghost Robotics Minitaur\cite{galloway_x-rhex_2010, kenneally_design_2016}. 

\section*{Acknowledgments}
This work was supported in part by the US Army Research Office under grant W911NF-17-1-0229. We thank J. Diego Caporale for his support with the experimental setup, Wei-Hsi Chen for his support with scaling, and  Charity Payne, Diedra Krieger, and the rest of the GRASP Lab staff  for keeping our lab safe and open during these tumultuous times. 
\bibliographystyle{ieee}

\bibliography{Jerboa_Initial_Lit_Review}
\newpage
\appendices
\section{Return Map Appendices}
\subsection{Maps}

\begin{table}[h]\centering
\begin{tabular}{|l|l|l|}
\hline
Name   & Definition & Ref \\ \hline
Descent map     &{$\!\begin{aligned} 
              \Psi_d :& \mathbb{R}^2  \rightarrow \mathbb{R}^3 \\   
               &[\dot{x}, y]_\text{apex} \mapsto [\dot{x}, y, \dot{y}]_{td} \end{aligned}$} & app. \ref{TR-decent_map}
\\ \hline
\pbox{20cm}{Flight to stance \\ reset map}    &{$\!\begin{aligned} 
              ^sR^f :& \mathbb{R}^3 \rightarrow \mathbb{R}^3 \times S^1 \\   
               &[\dot{x}, y, \dot{y}]_\text{td} \mapsto [r, \dot{r}, \theta, \dot{\theta}]_{td} \end{aligned}$}         
               & eq. \ref{TR-stance_to_flight}  \\ \hline
Stance map &{$\!\begin{aligned} 
              \Psi_s :& \mathbb{R}^3 \times S^1 \rightarrow \mathbb{R}^3 \times S^1 \\   
               &[r, \dot{r}, \theta, \dot{\theta}]_{td} \mapsto [r, \dot{r}, \theta, \dot{\theta}]_{lo} \end{aligned}$}       &sec. \ref{TR-stance_map}     \\ \hline
\pbox{20cm}{Stance to flight \\ reset map}&{$\!\begin{aligned} 
              ^fR^s :& \mathbb{R}^3 \times S^1  \rightarrow \mathbb{R}^3 \\   
               &[r, \dot{r}, \theta, \dot{\theta}]_{lo} \mapsto [\dot{x}, y, \dot{y}]_\text{lo} \end{aligned}$}& eq. \ref{TR-stance_to_flight}          \\ \hline
Ascent map    &{$\!\begin{aligned} 
              \Psi_a :& \mathbb{R}^3  \rightarrow \mathbb{R}^2  \\   
               &[\dot{x}, y, \dot{y}]_\text{lo} \mapsto [\dot{x}, y]_\text{apex} \end{aligned}$}    & app. \ref{TR-ascent_map}           \\ \hline
\end{tabular}
\caption{The maps that make up the apex coordinates return map for SLIP with attitude.}
\label{TR-maps}
\end{table} 

\subsection{Stance Dynamics}
After assumptions \ref{TR-ass_mom}, \ref{TR-ass_grav} the stance dynamics are

\begin{align}
    \begin{bmatrix}
    \ddot{r} \\
    \dot{\theta}
    \end{bmatrix} = \begin{bmatrix}
     \frac{\bar{p_\theta}^2}{m^2r^3}-\frac{k}{m}(r-r_g)-\frac{b}{m}\dot{r}  \\
     \bar{p_\theta}/(mr^2)
    \end{bmatrix}  \label{TR-post_assum_dyn}
\end{align}

\subsection{Stance Mode ODE Solution}
\label{TR-appendix:stanceFlow}
\subsubsection{Radial ODE Solution}

The fully simplified radial dynamics (\ref{TR-taylor}) are
\begin{equation}
    \ddot{r}=\frac{\bar{p}_\theta^2}{m^2r_g^3}-(\frac{3\bar{p}_\theta^2}{m^2r_g^4}+\frac{k}{m})(r-r_g)-\frac{b}{m}\dot{r} \label{TR-ddr_taylor2}.
\end{equation}
In order to solve equation \ref{TR-ddr_taylor2} in a more compact form let $\omega:=\sqrt{k/m+3\bar{p}_\theta^2/(m^2r_g^4)}$, let $\Gamma:=\bar{p}_\theta^2/(m^2r_g^3)+\omega^2r_g$, and let $\zeta:=b/(2m\omega)$. With these variables, equation \ref{TR-ddr_taylor2} can be written as
\begin{equation}
    \ddot{r}+2\zeta\omega\dot{r}+\omega^2r=\Gamma \label{TR-ddr_full_simple}
\end{equation}
Assuming $\zeta<1$, the solution to equation $\ref{TR-ddr_full_simple}$ is of the familiar form of a forced spring mass damper.

\begin{equation}
    r(t)=e^{-\zeta\omega t}(A\cos(\omega_d t)+B\sin(\omega_d t))+\Gamma/\omega^2 \nonumber
\end{equation}

Where $\omega_d:=\omega\sqrt{1-\zeta^2}$ and $A$ and $B$ are determined by the touchdown states, $r_{td}$ and $\dot{r}_{td}$
\begin{align}
    A&=r_{td}-\Gamma/\omega^2 \label{TR-rtd_def}  \nonumber\\
    B&=(\dot{r}_{td}+\zeta\omega A)/\omega_d \nonumber
\end{align}

We further simplify the radial flow with $M:=\sqrt{A^2+B^2}$ and $\psi:=\arctan2(-B,A)$ giving
\begin{equation}
    r(t)=Me^{-\zeta\omega t}\cos(\omega_d t+\psi)+\Gamma/\omega^2 \label{TR-r_final}
\end{equation}
Differentiation yields the radial velocity
\begin{equation}
    \dot{r}(t)=-M\omega e^{-\zeta\omega t}\cos(\omega_d t+\psi+\psi_2) \label{TR-dr_final}
\end{equation}
Where $\psi_2:=\arctan2(-\sqrt{1-\zeta^2},\zeta)$.

\subsubsection{Leg Angle ODE Solution}
With the analytical approximation for the radial dynamics, we can solve for the leg angle solution. Since angular momentum is conserved, $\dot{\theta}=\bar{p}_\theta/(mr^2)$. As with the radial dynamics, approximating $1/r^2$ with a Taylor series about $r=r_g$ yields a closed form solution.
\begin{equation}
    1/r^2 \approx 1/r_g^2-2/r_g^3(r-r_g)+\mathcal{O}((r-r_g)^2) \label{TR-taylor2}
\end{equation}

Using equation \ref{TR-taylor2} along with the radial solution yields
\begin{equation}
    \dot{\theta}=\frac{\bar{p}_\theta}{mr_g^2}(3-2\frac{M}{r_g}e^{-\zeta\omega t}\cos(\omega_d t+\psi)-2\frac{\Gamma}{r_g\omega^2}) \label{TR-dq}
\end{equation}

Integrating equation \ref{TR-dq} gives the leg angle trajectory
\begin{equation}
    \theta(t)=\theta_{td}+Xt+Y(e^{-\zeta\omega t}\cos(\omega_d t+\psi-\psi_2)-\cos(\psi-\psi_2)) \label{TR-theta_final}
\end{equation}
Where $X:=\frac{\bar{p}_\theta}{mr_g^2}(3-2\frac{\Gamma}{r_g\omega^2})$, and $Y:=\frac{2\bar{p}_\theta M}{mr_g^3\omega}$.

\subsection{Time of Liftoff Equation}
\label{TR-appendix:liftoff}

Liftoff is defined as the time when the force in the leg, $k(r-r_0)+b \dot{r} = 0$.
As in \cite{ankarali_stride--stride_2010} we assume the compression time is roughly equal to the decompression time. Thus $e^{-\zeta \omega t_{lo}}\approx e^{-\zeta \omega 2t_{b}}$.

We find $t_{b}$, the bottom time, by setting equation $\ref{TR-dr_final}$ equal to zero and solving.

\begin{equation}
    t_{b}=\frac{(2n_1+1)\pi/2-\psi-\psi_2}{\omega_d}, \, n_1\in\mathbb{Z} \nonumber
\end{equation}

Solving for $t_{lo}$ in $k(r-r_0)+b\dot{r}=0$ with the solutions to the dynamics substituted yields
\begin{equation}
    t_{lo}\approx \frac{2 n_2 \pi\pm\arccos(\frac{k(r_0\omega^2-\Gamma}{M_2M\omega^2 e^{-\zeta\omega2t_{b}}})-\psi-\psi_4}{\omega_d}, \, n_2\in\mathbb{Z} \label{TR-tlo} \nonumber
\end{equation}
Where $M_2:=\sqrt{k^2+b^2\omega^2-2bk\omega\cos(\psi_2)}$. In our operating regime of interest it is safe to take $n_1=0$, $n_2=1$, and the $\pm = -$.

\subsection{Ascent and Descent Map Derivations}
\label{TR-appendix:flight}
\subsubsection{Flight Trajectory}
In flight, the robot follows a purely ballistic trajectory about the center of mass. We describe this trajectory using cartesian coordinate.
\begin{align}
    \vthre{\dot{x}(t)}{y(t)}{ \dot{y} (t)} =
    \vthre{\dot{x}_0}{y_0+\dot{y}_0t-1/2 g t^2}{\dot{y}_0-g t} \label{TR-flight_flow}
\end{align}

\subsubsection{Ascent Map}
\label{TR-ascent_map}
Apex is defined as the state where $\dot{y} = 0$, thus $t_\text{apex} = \dot{y}_{lo}/g$. From here the ascent map, $\Psi_a$ is derived from equations \ref{TR-flight_flow} evaluated at $t = t_\text{apex}$. 

\subsubsection{Descent Map}
\label{TR-decent_map}
Touchdown occurs when the toe comes in contact with the ground. This is described by the equation $y(t) = r_0\cos(\theta_{td})$. Plugging in the flight solution gives
\begin{equation}
    y_{\text{apex}}+\dot{y}_{\text{apex}}t-1/2 g t^2 = \cos(\theta_{td})r_0 \label{TR-time_touchdown_simple} \nonumber
\end{equation}
Whose solution is
\begin{equation}
    t_{td} = \frac{\dot{y}_{\text{apex}}+\sqrt{\dot{y}_{\text{apex}}^2+2 g y_{\text{apex}}-2 g r_0 \cos(\theta_{td}) }}{g}
\end{equation}

From here the descent map, $\Psi_d$ is derived from equations \ref{TR-flight_flow} evaluated at $t = t_{td}$. 

\subsection{Reset Map Derivation}
\label{TR-appendix:resest}
Let $^fR^s$ be the reset map that maps from the stance state to the flight state at liftoff. 
  \begin{equation}
     z^f_{lo}=^fR^s(z_{lo}^s)=\begin{bmatrix}
-\dot{\theta}_{lo}r_{lo}\cos(\theta_{lo}) - \dot{r}_{lo}\sin(\theta_{lo})  \\
r_{lo} \cos(\theta_{lo}) \\
-\dot{\theta}_{lo}r_{lo}\sin(\theta_{lo}) + \dot{r}_{lo}\cos(\theta_{lo})\\
\end{bmatrix} 
\label{TR-stance_to_flight}
 \end{equation}

 Where, $z^f_{lo} = [\dot{x}, y, \dot{y}]_{lo}$ and $z_{lo}^s$ is the liftoff state in stance coordinates. 
 
Let  $^sR^f$ be the reset map that maps from the stance state to the flight state. This map changes coordinates from cartesian to polar.
\begin{equation}
    z_{td}^s=^sR^f(z_{td}^f) = \begin{bmatrix}
    r_0 \\
    -\sin (\theta_{td} ) \dot{x}+\cos (\theta_{td} ) \dot{y}    \\
    \theta_{td} \\
    \frac{-\cos (\theta_{td} ) \dot{x}-\sin (\theta_{td} ) \dot{y}}{r_0}\\

\end{bmatrix} 
\label{TR-flight_to_stance}
\end{equation}
Where $z^s_{td} = [r, \dot{r}, \theta, \dot{\theta}]_{td}$. 

\subsection{Approximate Solution of Angle of Attack}
 Starting from equation \ref{TR-eq:aoa} we use the Taylor series expansion of $\arctan(z) = \pi/4 z$ and the small angle approximation of $\cos{\theta} = 1- \theta^2/2$. We approximate the angle of attack as

\begin{equation}
    \theta_\text{AoA} \approx \frac{\pi}{4}(\frac{\dot{x}}{\sqrt{2 Ev/m - 2 g r_0 (1-k_\theta^2\theta_\text{AoA}^2/2)}}) \label{TR-eq:aoaApprox}.
\end{equation}

Equation \ref{TR-eq:aoaApprox} is quadratic in $\theta_\text{AoA}^2$, with coefficients
\begin{align*}
    a_\text{AoA} = &16 g r_0 \\
    b_\text{AoA} = &16(2 E_v/m - 2 g r_0) \\
    c_\text{AoA} = &-\dot{x}^2\pi^2
\end{align*}

Finally, we map the solution of (\ref{TR-eq:aoaApprox}) through $\Phi$ (\ref{TR-eq:aoa}) giving
\begin{equation}
    \theta_\text{AoA} \approx \Phi (\sqrt{Q_+(a_\text{AoA},b_\text{AoA},c_\text{AoA}}) \label{TR-Appendix:aoa_approx}
\end{equation}
 At the fixed points the maximum error between the analytical approximation for $\theta_\text{AoA}$ and the numerical solution to equation \ref{TR-eq:aoa} is $0.12$ rad  while the mean error is $0.03$ rad. .

\section{Details on Angle of Attack Fixed Points}
\label{TR-app:aoa}
This appendix should be used in conjunction with section \ref{TR-sec:fixed}.

\subsection{Constants in Simplified Stance Map}
The constants in equation \ref{TR-simple_stance} are defined as
\begin{align}
    C_1 := & \frac{\omega e^{-\zeta t_{lo} \omega} \left(\sqrt{1-\zeta^2} \cos (t_{lo} \omega_d)-\zeta \sin (t_{lo} \omega_d)\right)}{\omega_d}\label{TR-C1} \\
    C_2 := & \frac{A\omega e^{-\zeta  \omega t_{lo}}}{\omega_d} \left(- \sqrt{1-\zeta^2}  \omega_d \sin (  \omega_d t_{lo}) \right. \nonumber \\ 
    &\left.  -\zeta^2 \omega \sin ( \omega_d t_{lo}) + \zeta \cos ( \omega_d t_{lo}) \left(\sqrt{1-\zeta^2} \omega- \omega_d\right)\right)  \\
    C_3 :=& \frac{2 \sqrt{1-\zeta^2} p_\theta e^{-\zeta \omega t_{lo}} \cos (\omega_d t_{lo})}{m r_g^3 \omega \omega_d} \nonumber \\
    &+\frac{2 \zeta p_\theta e^{-\zeta \omega t_{lo}} \sin (\omega_d t_{lo})}{m r_g^3 \omega \omega_d} \nonumber\\
    &-\frac{2 \sqrt{1-\zeta^2} p_\theta}{m r_g^3 \omega \omega_d} \\
    C_4 :=& -\frac{2 A \sqrt{1-\zeta^2} \zeta p_\theta}{m r_g^3 \omega_d} +\frac{2 A \zeta^2 p_\theta e^{-\zeta\omega t_{lo}} \sin p_\theta)}{m r_g^3 \omega_d} \nonumber \\
    & -\frac{2 A \sqrt{1-\zeta^2} p_\theta e^{-\zeta\omega t_{lo}} \sin p_\theta)}{m r_g^3 w} -\frac{2 A \zeta p_\theta}{m r_g^3 w} \nonumber \\
    &+\frac{2 A \sqrt{1-\zeta^2} \zeta p_\theta e^{-\zeta\omega t_{lo}} \cos p_\theta)}{m r_g^3 \omega_d} \nonumber \\
    &+\frac{2 A \zeta p_\theta e^{-\zeta\omega t_{lo}} \cos p_\theta)}{m r_g^3 w}-\frac{p_\theta t_{lo} \left(2 \Gamma-3 r_g w^2\right)}{m r_g^3 w^2} \label{TR-C4}
\end{align}

\subsection{Derivation of Assumption \ref{TR-ass_aoa}}
$\theta_{td}:=k_\theta\theta_\text{AoA} = \theta_{AoA}-\theta_\text{AoA}(1-k_\theta)$. Assuming $\theta_{AoA}$ is nominally $\Bar{p}_\theta/0.7\frac{\pi}{4}$, then $\theta_{td} \approx \theta_\text{AoA} - \theta_\text{offset}$ where $\theta_\text{offset} := \Bar{p}_\theta/0.7\frac{\pi}{4}(1-k_\theta)$. From trigonometry, we get that for a given $k_\theta$, $\dot{r}_{td} \approx -||v_{td}|| \cos{(\theta_\text{offset})}$ and $\dot{\theta}_{td} \approx ||v_{td}||/r_0 \sin{(\theta_\text{offset})} = -\dot{r}_{td}/r_0\tan{(\theta_\text{offset})}$ where $||v_{td}|| := \sqrt{\dot{x}^2_{td}+\dot{y}^2_{td}}$.
\label{TR-Appendix_aoa_assumption}

\subsection{Solving For The Fixed Points}
After  assumption \ref{TR-ass_aoa} the constraints on energy and speed are
\begin{align}
    E_{td} &= 1/2 m \Dot{r}^*_{td}{}^2 + 1/2 m  (\Dot{r}^*_{td} \tan\theta_\text{offset})^2 = \nonumber \\
    E_{lo} & = 1/2 m (C_1 \Dot{r}^*_{td}+C_2)^2 + \frac{\Bar{p}_\theta^2}{2 m r_0^2} \label{TR-e1}\\
    \Dot{x}_{td} & = -\Dot{r}^*_{td}\theta^*_{td}+\Dot{r}^*_{td} \tan\theta_\text{offset} (1- \theta^*_{td}{}^2/2)  = \nonumber\\
    \Dot{x}_{lo} & = \frac{-\Bar{p}_\theta}{m r_0}(1-(\theta^*_{td}+C_3 \Dot{r}^*_{td}+C_4)^2/2 )\nonumber \\&-(C_1 \Dot{r}^*_{td}+C_2) (\theta^*_{td}+C_3 \Dot{r}^*_{td}+C_4). \label{TR-dx2}
\end{align}

The constrain on energy is quadratic in $\dot{r}^*_{td}$ with coefficients
\begin{align}
    \vthre{ a_r(\args)}{ b_r}{ c_r(\bar{p}_\theta )} := \vthre{m/2 (1-C_1^2 +\tan^2(\theta_\text{offset}))}{-C_1 C_2 m}{-\frac{\bar{p}^2_\theta}{2 m r_0^2} - \frac{C_1^2 m}{2}}
    \label{TR-r_coeff}
\end{align}

Similarly the constraint on speed is quadratic in $\theta^*_{td}$ with coefficients
\begin{align}
    a_\theta(r^*_{td},\args)  := &1/2 \dot{r}^*_{td} \tan(\theta_\text{offset}) - \frac{\bar{p}_\theta}{2 m r_0}  \label{TR-a_theta}\\
    b_\theta(r^*_{td}, \args)  :=&-\dot{r}^*_{td} - (\frac{C_4 \bar{p}_\theta + C_3 \dot{r}^*_{td} \bar{p}_\theta}{m r_0} - C_2 - C_1 \dot{r}^*_{td}) \\
    c_\theta(\dot{r}^*_{td}, \args)  := &(C_2 + C_1 \dot{r}^*_{td}) (C_4 + C_3 \dot{r}^*_{td}) + \nonumber\\
    &\dot{r}^*_{td}\tan(\theta_\text{offset}) - \frac{(-2 + (C_4 + C_3 \dot{r}^*_{td})^2)\bar{p}_\theta}{2 m r_0}\label{TR-b_theta}
\end{align}

The quadratic root function is
\begin{align}
    \begin{bmatrix}
     Q_+(a,b,c) \\
      Q_-(a,b,c) \\
    \end{bmatrix} = \begin{bmatrix}
    \frac{-b + \sqrt{b^2 - 4 a c}}{2a} \\
    \frac{-b - \sqrt{b^2 - 4 a c}}{2a}
    \end{bmatrix}
    \label{TR-quadratic_root}
\end{align}
\section{Fixed Point Accuracy}
\begin{figure*}
     \centering
     \begin{subfigure}[b]{ \textwidth}
         \centering
         \includegraphics[width=\textwidth]{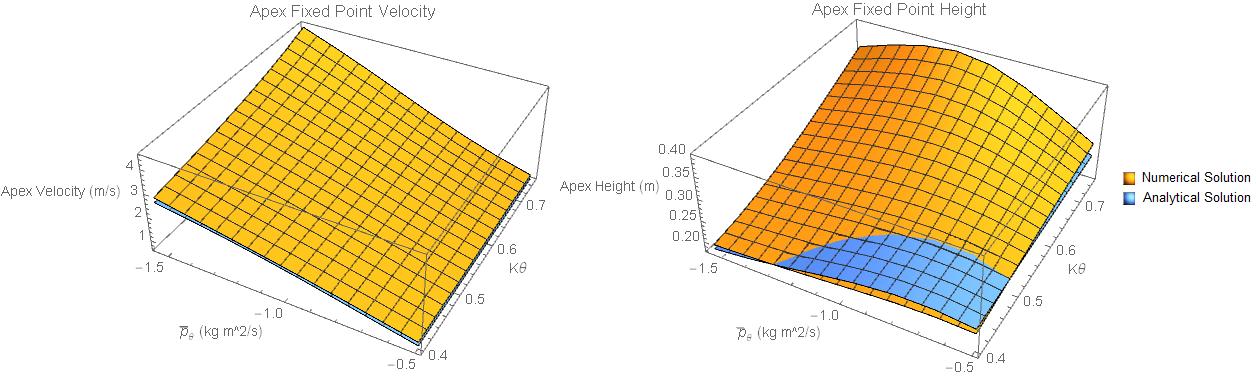}
         \caption{Apex coordinate fixed points for the approximate analytical return map and the numerical return map over a rectangular approximation of the robot's operating regime. All of the fixed points are stable.}
     \end{subfigure}
     \begin{subfigure}[b]{\textwidth}
         \centering
         \includegraphics[width=\textwidth]{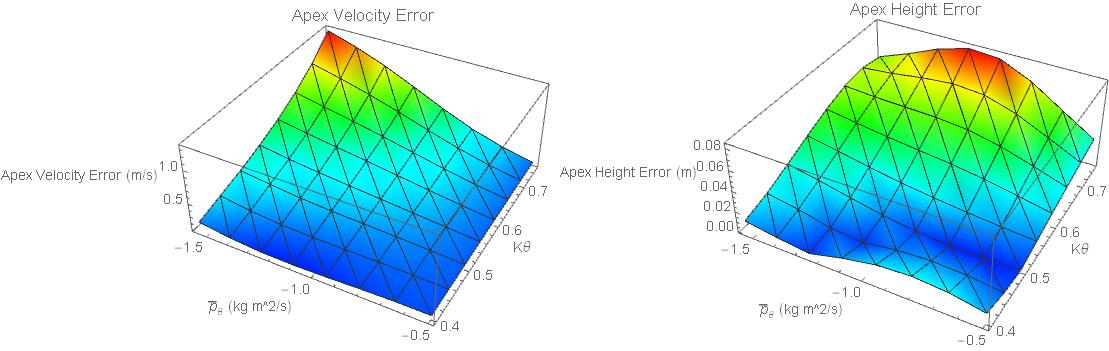}
         \caption {Error between apex coordinate fixed points for the approximate analytical return map  and the numerical return map  over a rectangular approximation of the robot's operating regime.}
     \end{subfigure}
        \caption{ Accuracy of the apex coordinate fixed points over a large set of gains, $\Bar{p}_\theta \in [-0.5, -1.55] kg m^2/s$, $k_\theta \in [0.4, 0.75]$. The SLIP apex coordinates, $\dot{x}$ and $y$, show that approximation start to break down for the higher $\Bar{p}_\theta$ and $k_\theta$. The analytical map underestimates the speed. The model parameters for these fixed points are in table \ref{TR-parameters}.}
        \label{TR-fig:fixedPointLargeError}
\end{figure*}

Figure \ref{TR-fig:fixedPointLargeError} has the error of the fixed points over a rectangular approximation of the robot's operating regime.

\end{document}